\pdfoutput=1

\documentclass[11pt]{article}

\usepackage[]{acl}

\usepackage{times}
\usepackage{latexsym}
\usepackage{amsmath}
\usepackage{tikz}
\usepackage{algorithm}
\usepackage{algorithmic}
\usepackage{booktabs}

\usepackage[T1]{fontenc}
\usepackage{graphicx}

\usepackage[utf8]{inputenc}

\usepackage{microtype}

\usepackage{inconsolata}

\usepackage{hyperref}

\makeatletter
\def\thanks#1{\protected@xdef\@thanks{\@thanks
        \protect\footnotetext{#1}}}
\makeatother

\title{Numerical Claim Detection in Finance: A New Financial Dataset, Weak-Supervision Model, and Market Analysis}

\author{\hypersetup{linkcolor=black} Agam Shah$^{\heartsuit\  \dag}$, Arnav Hiray$^{\heartsuit \ \dag}$\;, Pratvi Shah$^{\clubsuit}$, Arkaprabha Banerjee$^{\clubsuit}$, Anushka Singh$^{\diamondsuit}$\\ \bf Dheeraj Eidnani$^{\heartsuit}$, Sahasra Chava$^{\spadesuit}$, Bhaskar Chaudhury$^{\clubsuit}$, Sudheer Chava$^{\heartsuit}$\\
$^{\heartsuit}$ Georgia Institute of Technology\\
$^{\clubsuit}$ DA-IICT\\
$^{\diamondsuit}$  IIT-Kharagpur\\
$^{\spadesuit}$ Fulton Science Academy
\thanks{Correspondence to Agam Shah \textcolor{darkblue}{{\{\href{mailto:ashah482@gatech.edu}{ashah482@gatech.edu}\}}}
$\dag$ These authors contributed equally to this work}}

\begin{document}
\maketitle
\begin{abstract}
In this paper, we investigate the influence of claims in analyst reports and earnings calls on financial market returns, considering them as significant quarterly events for publicly traded companies. To facilitate a comprehensive analysis, we construct a new financial dataset for the claim detection task in the financial domain. We benchmark various language models on this dataset and propose a novel weak-supervision model that incorporates the knowledge of subject matter experts (SMEs) in the aggregation function, outperforming existing approaches. We also demonstrate the practical utility of our proposed model by constructing a novel measure of \textit{optimism}. Here, we observe the dependence of earnings surprise and return on our optimism measure. Our dataset, models, and code are publicly (under CC BY 4.0 license) available on GitHub\footnote{ \url{https://github.com/gtfintechlab/fin-num-claim}.}. 
\end{abstract}

\section{Introduction}

Earnings conference calls are a quarterly event where the company's top executives provide performance reports of the company over the last quarter (3 months). Between the two earnings calls analyst from various financial institutions analyze and provide earnings estimates and recommendations. For example, \citet{jegadeesh2010analysts} has documented that there is a significant stock market reaction to analysts' recommendations (ratings). Recent insights, such as those presented by \citet{mclean2020retail}, reveal that retail investors, often perceived as unsophisticated, exhibit responsiveness to analysts' projections, underscoring the pivotal role of analysts' reports in informing market participants.  However, analyst ratings can be biased \citep{michaely1999conflict, corwin2017investment, coleman2021human}. Therefore it is important to understand whether the ratings are backed by strong numerical financial claims in the analyst's report. Further, the sentences with a claim have a higher density of forward-looking information. As an application, extraction of numerical ESG claims from earnings call transcripts, can help better understand whether companies do walk the talk on their environment and social responsibility claims \citep{chava2021ESG}. These examples underscore the necessity of numerical claim detection in the finance domain, aligning with broader research efforts to ensure the accuracy and reliability of information sources.

\begin{figure*}[!tbp]
  \centering
  \begin{minipage}[b]{0.37\textwidth}
    \includegraphics[width=\textwidth]{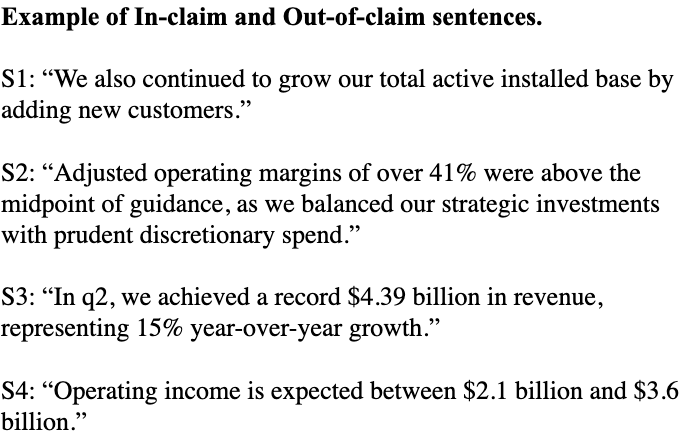}
  \end{minipage}
  \hfill
  \begin{minipage}[b]{0.62\textwidth}
    \includegraphics[width=\textwidth]{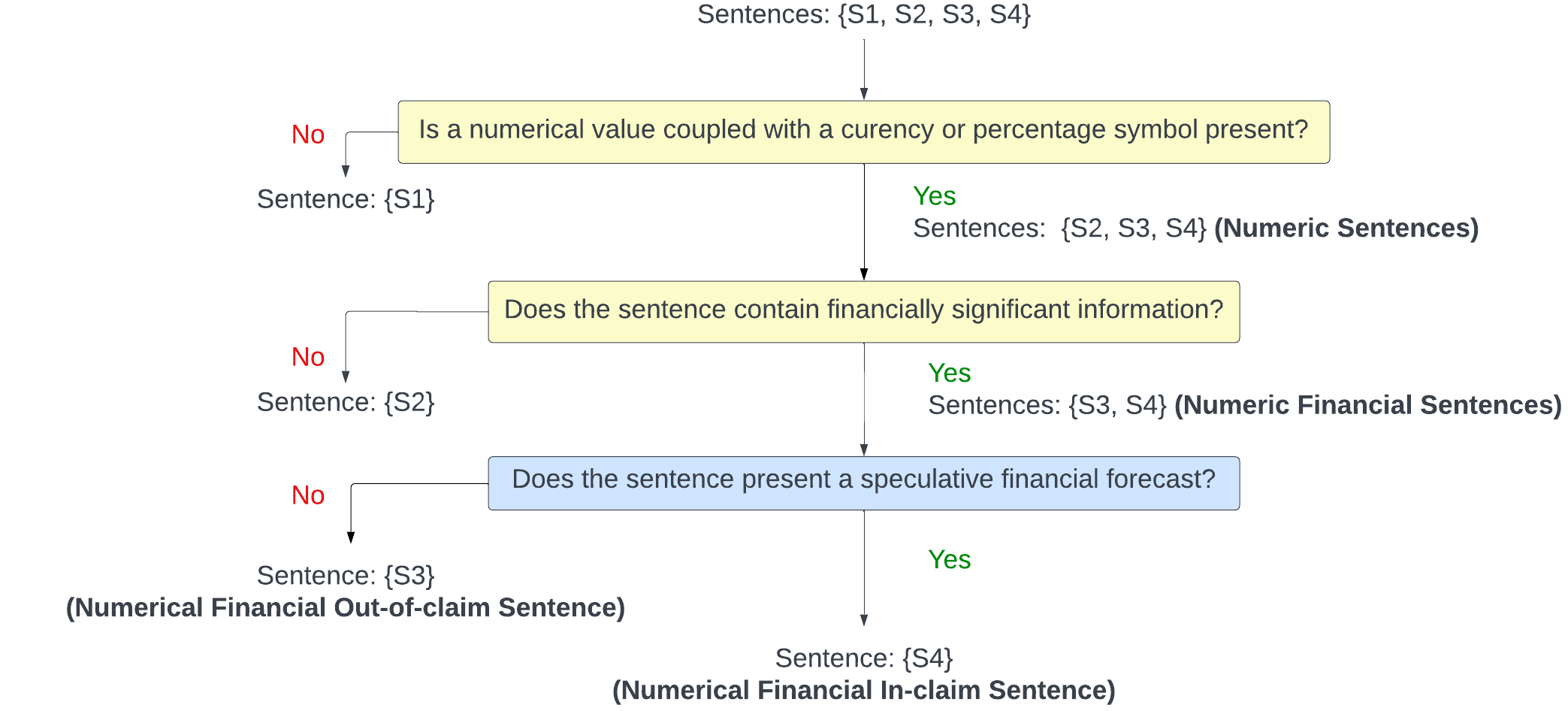}
  \end{minipage}
  \caption{Example of In-claim and Out-of-claim sentences.}
  \label{fig:examples} 
\end{figure*}

A key component of this paper is the identification of Numeric Financial Sentences. Specifically, Numeric Financial Sentences include a financial term, a numeric value, and either a currency or percentage symbol. \citet{chen2020numclaim} first introduced the categorization of sentences into `in-claim' and `out-of-claim' specifically in the Mandarin language. Expanding on their foundation, we define an `in-claim' sentence as one presenting a speculative financial forecast. Conversely, an `out-of-claim' sentence presents a numerical statement about a past event, transitioning from a mere claim to a confirmed fact.  For clarity, `in-claim' sentences can also be termed "financial forecasts" whereas `out-of-claim' can be labeled as "established financials." Every Numeric Financial Sentence that is not a speculative financial forecast (in-claim) is then identified as an `out-of-claim' sentence. Figure \ref{fig:examples} illustrates the identification of Numeric Financial Sentences as well as distinguishing between ``in-claim" and ``out-of-claim" sentences.

A major challenge for building or training predictive models is the scarcity of labeled data \citep{zhang2021wrench, ratner2017snorkel}. Supervised learning often involves a significant amount of manual labeling of data which is often infeasible for large datasets. In such scenarios, one can leverage weak-supervision-based learning methods \cite{varma2018snuba} or fine-tune the pre-trained language model. Weak-supervision is a process that leverages slightly noisy or imprecise labeling functions (lfs) to label vast amounts of unlabeled data \cite{ratner2020snorkel, lison2021skweak}. The strength of the weak-supervision model lies in these imperfect labels, when combined, producing improved predictive models \cite{lison2021skweak, zhang2021wrench}. However, a crucial component involves the development of effective lfs for a given raw dataset systematically rather than manual annotation \cite{lison2021skweak}. 

The aim of our work is to derive financially significant information from the quarterly analyst reports and earnings calls by categorizing each numerical sentence as in-claim or out-of-claim. Our major contributions through this paper are the following:
\begin{itemize}
    \itemsep0em 
    \item We introduce a new task of claim detection (in English) with a labeled dataset.
    \item We build clean, tokenized, and annotated open-source datasets based on earnings calls. 
    \item We introduce a weak-supervision model with a novel aggregation function. 
    \item We benchmark a wide range of language models for the claim detection task. 
    \item We develop a novel measure of optimism and validate its usefulness in predicting various financial indicators. 
\end{itemize}

\section{Related Work}
\paragraph{NLP in Finance} Finance is one of the most attractive domains for the application of NLP. \citet{araci2019finbert} and \citet{liu2020finbert} presented pre-trained language models for the Finance domain. There are multiple datasets specifically catered for applications of NLP in finance including question answering dataset created by \citet{chen2021finqa} and \citet{FiQA10.1145/3184558.3192301}, and also a NER dataset constructed by \citet{shah2023finer} for the financial domain. There is a vast body of literature on undertaking sentiment analysis tasks on financial data\citep{FiQA10.1145/3184558.3192301, malo2014good, day2016deep, akhtar2017multilayer}. 

Works of \citet{li2020maec} and \citet{sawhney2020voltage} were centered around predicting volatility using earnings call transcripts in the domain of risk management. \citet{chava2022measuring} measure the firm level inflation exposure by fine-tuning RoBERTa \citep{roberta}, while \citet{li2021measuring} leveraged word-embeddings to measure the corporate culture. Moreover, \citet{nguyen2021multimodal} and \citet{daltonpersuading} used multimodal machine learning for credit rating prediction and measurement of persuasiveness respectively. \citet{shah2023trillion} investigated the impact of monetary policy communication on financial markets. \citet{cao2020talk} critically examined the evolution of corporate disclosure in recent years, influenced by the rising application of NLP in Finance. Our research focuses on identifying numerical financial claims from a vast set of English analyst reports and earnings calls using a weak-supervision model. This differs from \citet{chen2020numclaim}, which targets numeric claim detection in a smaller Chinese language dataset.

\paragraph{Weak-Supervision} In order to reduce the complexities associated with manual labeling, several standard techniques such as semi-supervised learning \cite{chapelle2009semi}, transfer learning \cite{5288526}, and active learning \cite{settles2009active} have been employed. However, many researchers \citep{meng2018weakly, kartchner2020regal} and practitioners also employ weak-supervision-based models to further reduce the computational costs while retaining the accuracy of the labeled data. Weak-supervision models were primarily developed in a bid to replace standard labeling techniques with models which can leverage slightly noisy or imprecise sources to label vast amounts of data \cite{ratner2020snorkel}. Techniques such as distant supervision \cite{mintz2009distant} and crowd-sourced labels \cite{yuen2011survey} are often associatewd with weak-supervision-based models, however, they tend to have limited coverage and accuracy \cite{lison2021skweak}. In the case where we have noisy labels from multiple sources available, there have been efforts made to use majority vote, weighted majority vote \citep{ratner2020snorkel}, and other label-models \citep{yu2022actune, zhang2022prboost}.

\section{Dataset}
\label{section_datatset}
We collect two categories of text and financial market datasets. Analyst reports are procured from a proprietary source while earnings call transcripts are collected in a manner that allows us to make the resulting dataset open-source. 

\subsection{Analyst Reports}
The raw dataset consists of quarterly analyst reports (in English) for a large number of public firms in the U.S. These analyst reports were collected from Zacks Equity Research and were available to us through the Nexis Uni license\footnote{Nexis Uni license doesn't authorize republication of full or partial text. To solve this problem, we also collect and construct a dataset from earnings calls which can be made public under CC BY 4.0 license.}. 

The text documents are first split into sentences using multiple regex-based rules. This segmentation process utilizes a comprehensive set of regular expression (regex) rules to accurately identify sentence boundaries,  accounting for a variety of English language nuances, including abbreviations, titles, websites, and numerical expressions, to ensure precise sentence delineation.  We employ regex-based rules as they typically are significantly faster with similar accuracy compared to standard libraries in sentence tokenization. Next, sentences containing quantitative data - specifically sentences with a numeric value AND either a currency symbol as a prefix or percentage symbol as a postfix- are extracted, as they have numerical relevance \cite{chen2019numeral}. This numerical condition filter reduced the number of sentences by 66.7\%.

The next step in the pipeline uses a whitelisting technique to retain only sentences with financially significant information, achieved by cross-referencing each sentence with a financial dictionary containing a comprehensive list of financial market terms and related literature. The financial dictionary used in this study, developed by \citet{shah-etal-2022-flue}, contains over 8,200 financially significant terms. Sentences are cross-referenced with this dictionary to verify financial significance; if no words match, the sentence is marked as irrelevant. This filtering reduced the dataset by an additional 17.2\%. The dataset contains 8,583,093 total sentences, 2,857,567 numeric sentences, and 2,364,977 numeric-financial sentences after filtration. This two-tier filtering method enriched the data by retaining only 27.5\% of the sentences from the original data.

\subsection{Earnings Call Transcripts}
\label{section_datatset_EC}To make our work more impactful, we also collect earnings call transcripts for NASDAQ 100 companies from their investor relation page. We were able to write individual scripts for 78 out of 100 NASDAQ companies. As all the companies in this list are public companies, their data can be accessed and shared publicly which allows us to open-source the resulting dataset. Collecting data till March of 2023 results in a total of 1,085 earnings call transcripts. The biggest advantage of writing separate scripts for each company is that it allows us to keep adding more transcripts every quarter increasing the size of the dataset shared over time. We apply text processing (tokenization, numerical filter, financial dictionary filter) on earnings call transcripts similar to what is used for analyst reports.

\begin{table*}[ht]
\centering
\footnotesize
\begin{tabular}{lccc}
\toprule
\textbf{Dataset} & \textbf{Analyst Reports} & \textbf{Earnings Calls} & \textbf{NumClaim \citep{chen2020numclaim}} \\
\midrule
Language & English & English & Chinese \\
Year & 2017-20 & 2017-23 & NA \\
Sector Information & Yes & Yes & No \\
\# Stocks & 1,530 & 78 & NA \\
\# Files & 87,536 & 1,085 & NA \\
\# Words & 167,301,873 & 11,641,673 & 42,594 \\
\# Numeric Sentences & 2,857,567 & 48,686 & 5,144 \\
\# Numeric Financial Sentences & 2,364,977 & 41,013 & NA \\
\# Numeric Financial In-Claim Sentences & 336,252 & 5362 & 1,233 \\
\bottomrule
\end{tabular}

\caption{Comparison of our datasets with NumClaim \citep{chen2020numclaim} dataset. }
\label{tb:dataset_comparision}
\end{table*}

\subsection{Comparison with Related Dataset}
In this section we compare our proposed datasets with NumClaim \cite{chen2020numclaim}, an expert-annotated dataset in the Chinese language. Our dataset of raw analyst reports in the English Language from 1,530 major companies over the period of 2017-20 is significantly larger than NumClaim or other associated datasets. Our open-sourced dataset from collected earnings call transcripts is also larger than the NumClaim dataset. The detailed comparison of our datasets with NumClaim is provided in Table \ref{tb:dataset_comparision}.

\subsection{Financial Market Data}
\paragraph{Stock Price and Earnings Surprise Data}
We collect stock price data from Polygon.io\footnote{\url{https://polygon.io/stocks}} starting January 1st, 2017. We collect the actual earnings per share (EPS) and forecasted median EPS from the I/B/E/S dataset\footnote{\url{https://www.investopedia.com/terms/i/ibes.asp}}. 

\paragraph{Sector Data}
For each firm in our dataset, we collect sector information by collecting GSECTOR classification from the annual fundamental COMPUSTAT database. GSECTOR maps each company to one of the twelve sectors. 

\subsection{Sampling and Manual Annotation}
From the complete raw dataset of 87,536 analyst reports and 1,085 earnings call transcripts, we sample data and annotate sentences. The sampled dataset consisted of 96 analyst reports consisting of two files per sector per year, accounting for about 2,681 unique financial-numeric sentences. We also sample 12 earnings call transcripts randomly consisting of two files per year, consisting of 498 financial-numeric sentences. This set was manually annotated and assigned `in-claim' or `out-of-claim' labels by two of the authors with a foundational background in finance (one of them is now an analyst at a top investment bank) and domain expertise developed through examples provided by a co-author. This co-author is a financial expert with a Master's degree in Quantitative Finance, currently pursuing a PhD under the guidance of the Chair Professor of Finance, and has contributed to work at leading finance journals and conferences. The annotator agreement was 99.21\% and 95.78\% for analyst reports and earnings call transcripts respectively. Any disagreement between the two annotators was resolved with the help of the financial expert mentioned earlier. The dataset (Train, Val, Test) is split as follows: Analyst Reports (1,715, 429, 537) and Earnings Calls (318, 80, 100).

\section{Experiments}
\subsection{Models}
In this section, we provide details of the four categories of models we have used. Initially, we provide detail on the proposed weak-supervision model with the customized aggregation function. In order to provide a comprehensive benchmark for the claim detection task and comparison with proposed weak-supervision model, we add Bi-LSTM, six BERT architecture-based PLMs, and three generative LLMs.

\paragraph{Weak-Supervision Model}
For implementing a weak-supervision model we use the Snorkel library \cite{ratner2017snorkel}, leveraging its inherent pipeline structure for generating labels for each data segment and then passing the outputs through the customized aggregation function. 

Labeling functions used in our model include rule-based pattern matching combined with part-of-speech (POS) tag constraints for some phrases. We create seventeen labeling functions for the categorization of results and also make use of multiple other labeling functions in order to divide sentences representing assertions or written in the past tense. These labeling functions are listed in Table \ref{labeling_functions}. More details on the construction of the labeling function can be found in Appendix \ref{sec:ap_labeling_function}. 

\paragraph{Aggregation Function}The output of the labeling functions needs to be aggregated to decide the final label of the sentence. Unlike other models, we use independent, weighted labeling functions with weights based on the level of confidence assigned by Subject Matter Experts (SMEs). Our labeling function can produce four distinct types of output: -1 for a high confidence out-of-claim sentence, 0 for abstention from making a claim, 1 for a low confidence in making a claim, and 2 for a high confidence in making a claim. This system allows us to further differentiate in-claim sentences into two levels of confidence. The pseudo-code in Algorithm~\ref{alg:aggregation_function} illustrates our aggregation function.

\begin{algorithm}
\footnotesize
\begin{algorithmic}
\IF{any of the labeling functions' output is $-1$}
    \STATE $label \gets$ "out-of-claim"
\ELSIF{the max of the labeling functions' output is $2$}
    \STATE $label \gets$ "in-claim"
\ELSE
    \STATE $label \gets$ majority vote output
\ENDIF
\end{algorithmic}
\caption{Aggregation Function}
\label{alg:aggregation_function}
\end{algorithm}

Traditional majority vote takes decisions based on votes from all the labeling functions, meaning assigning equal weights. The weighted majority vote aggregation function, such as Snorkel, learns the weight for each labeling function from the data itself. In our case, Subject Matter Experts decide that some labeling functions are higher in the hierarchy than others. This means that we look at their labels first before looking at the output of other labeling functions. If those higher-valued labeling functions refrain from voting (by giving an abstain label, value=0), we look at the output of other labeling functions. Otherwise, we take labels based on the majority vote.

\begin{table*}[ht]
\centering
\begin{tabular}{lcccc}
  \toprule
  \multicolumn{5}{c}{\textbf{Panel A: Models Without Further Training}}\\
  \midrule
  \textbf{Model} & \multicolumn{2}{c}{\textbf{Analyst Reports (AR)}}  & \multicolumn{2}{c}{\textbf{Earnings Calls (EC)}}\\ 
  \midrule
  Weak-Supervision & \multicolumn{2}{c}{0.9272 (0.0116)}& \multicolumn{2}{c}{0.9382 (0.0213)}\\
  \midrule
  Falcon-7B (0-shot) & \multicolumn{2}{c}{0.4167 (0.0075)} & \multicolumn{2}{c}{0.3884 (0.0624)}\\
  Llama-2-70B (0-shot) & \multicolumn{2}{c}{0.7278 (0.0079)} & \multicolumn{2}{c}{0.5407 (0.0267)}\\
  ChatGPT-3.5 (0-shot) & \multicolumn{2}{c}{0.9191 (0.0144)} & \multicolumn{2}{c}{0.7569 (0.0023)}\\
  \midrule
  Falcon-7B (6-shots) & \multicolumn{2}{c}{0.3410 (0.0109)} & \multicolumn{2}{c}{0.3021 (0.0343)}\\
  Llama-2-70B (6-shots) & \multicolumn{2}{c}{0.9169 (0.0049)} & \multicolumn{2}{c}{0.7972 (0.0228)}\\
  ChatGPT-3.5 (6-shots) & \multicolumn{2}{c}{0.8943 (0.0033)} & \multicolumn{2}{c}{0.7334 (0.0198)}\\
  \bottomrule
  \toprule
  \multicolumn{5}{c}{\textbf{Panel B: Fine-Tuned Models}} \\
  \midrule
  \textbf{Train/Test}  & \textbf{AR/AR}  & \textbf{EC/AR} & \textbf{AR/EC}  & \textbf{EC/EC} \\ 
  \midrule
  Bi-LSTM & 0.9309 (0.0235) & 0.8244 (0.0332) & 0.8961 (0.0236) & 0.8892 (0.0375)\\
  \midrule
  BERT-base-uncased & 0.9532 (0.0192) & 0.9269 (0.0150) & 0.9251 (0.0113) & 0.9376 (0.0205)\\
  FinBERT-base & 0.9617 (0.0076) & 0.9381 (0.0112) & 0.9209 (0.0257) & 0.9279 (0.0135)\\
  FLANG-BERT-base & 0.9611 (0.0137) & 0.9270 (0.0109) & 0.9119 (0.0257) & 0.9363 (0.0089)\\
  RoBERTa-base & 0.9615 (0.0091) & 0.9319 (0.0131) & 0.8906 (0.0301) & \textbf{0.9563} (0.0036)\\
  \midrule
  BERT-large-uncased & 0.9539 (0.0111) & 0.9183 (0.0063) & 0.9197 (0.0349) & 0.9416 (0.0349)\\
  RoBERTa-large & \textbf{0.9642} (0.0069) & 0.9381 (0.0138) & 0.8975 (0.0244) & 0.9427 (0.0153)\\
   \bottomrule
\end{tabular}

\caption{In the table, A/B indicates that the model is fine-tuned on dataset A and tested on dataset B. All values are F1 scores. An average of 3 seeds was used for all models. The standard deviation of F1 scores is in parentheses. }
\label{tb:master_accuracy}
\end{table*}

To facilitate a comprehensive comparison of our weak-supervision model against various other model categories, we additionally leverage Generative Large Language Models (LLMs) in both zero-shot and few-shot settings, and conduct fine-tuning on Bi-LSTM as well as other Pre-trained Language Models (PLMs). Detailed information regarding the implementation of these models is delineated in the Appendix \ref{ap:model_details}.

\subsection{Results}
In this section, we present the results obtained using the above models and provide a detailed analysis of the outcomes. 
\paragraph{Weak-Supervision Model}
The performance in Table \ref{tb:master_accuracy}, highlights how well our Weak-Supervision based model performs when compared with manually annotated data. In order to make sure that there is no contamination issue between the labeling functions and annotated data, we perform a robustness check in Appendix \ref{sec:robustness}. We also perform ablation on the number of labeling functions in Appendix \ref{ap:number_lfs}.

We consider majority voting and Snorkel's aggregation function \cite{ratner2017snorkel} as baseline aggregation functions for comparative ablation analysis. The accuracy of baseline aggregation functions along with our aggregation function is reported in Table \ref{tb:baseline_aggregators}. For all three models, the same set of labeling functions is used and they only differ in the aggregation part.\footnote{We do not perform any post-processing on the output to convert abstain label to one of the labels.} The result highlights the importance of the construction of a customized aggregation function for a weak-supervision model where a small set of labeling functions are complete and less noisy.

\begin{table}[ht]
\centering
\footnotesize
\begin{tabular}{ccc}
\toprule
\textbf{Aggr. Funtion} & \textbf{AR}  & \textbf{EC}\\
\midrule
Majority Vote & 0.4274 (0.0208) & 0.5313 (0.0427)\\
Snorkel's WMV & 0.4269 (0.0204) & 0.5309 (0.0372)\\
Ours           & 0.9272 (0.0116) & 0.9382 (0.0213)\\
\bottomrule 
\end{tabular}
\caption{Performance comparison of our aggregation function with baseline aggregation functions. All values are F1 scores. An average of 3 seeds was used for all models. The standard deviation of F1 scores is reported in parentheses.}
\label{tb:baseline_aggregators}
\end{table}

\paragraph{Generative LLMs} There are a few observations regarding the performance of Generative LLMs. First, we see that utilizing a more detailed prompt leads to large improvements in performance across all three models. Secondly, Falcon and Llama have a large increase in performance as well when using six-shot prompting. However, ChatGPT did not have as large of an improvement when utilizing few-shot prompting. While the reasoning behind this is uncertain, it is clear that prompt engineering (particularly creating detailed prompts) can lead to substantial improvement.
Zero-shot ChatGPT fails to outperform both weak-supervision and fine-tuned PLMs. It still achieves impressive performance without having access to any labeled data. Of the variations of prompting attempted, Llama with six-shot prompting yielded the best results. This seems to suggest that through the use of prompt engineering, open-source models may be able to close the gap with closed LLMs. 

\paragraph{Bi-LSTM}
The Bi-LSTM model outperforms the weak-supervision model on analyst reports data but doesn't outperform on earnings call data. The potential reason can be the larger fine-tuning dataset available for analyst reports. It doesn't outperform the model based on BERT on any of the four configurations. 

\paragraph{PLMs}
The fine-tuned models utilizing the BERT architecture demonstrate superior performance compared to other model classes, emphasizing the significant value gained from annotated data. Intriguingly, the model that achieves the highest performance within a particular train-test dataset category does not necessarily exhibit the best performance on transfer learning datasets. This finding underscores the importance of separate data annotation. Notably, the RoBERTa model emerges as the top performer within the same train-test data category.

\begin{figure}
    \centering
    \includegraphics[width=\linewidth]{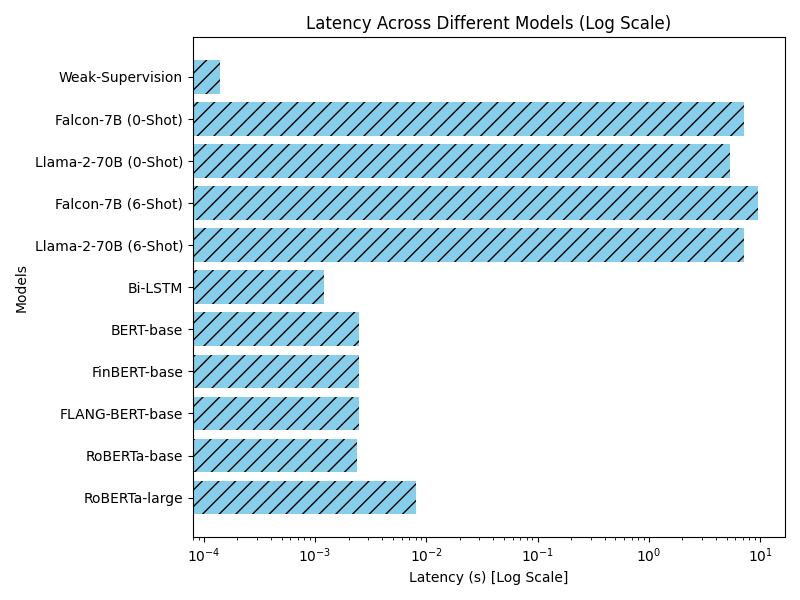}
    \caption{This bar chart compares the latency (log scale) of various models relative to the weak-supervision model.}
    \label{fig:latency}
\end{figure}

\paragraph{Latency and Financial Applicability}
In finance, latency is crucial as investors aim to surpass competitors. Figure~\ref{fig:latency} shows just how stark the differences is in latency. Our weak-supervision (WS) model stands out for its low latency, offering significant advantages in the fast-moving financial markets. Despite challenges in measuring latency for API-based, closed-source models like ChatGPT, our analysis on Falcon-7B and Llama-70B highlights the WS model's superior speed and efficiency. This model's performance is key in finance, where processing speed can be decisive in transaction success. Furthermore, even if generative LLMs do overcome the hurdle of latency, large ethical challenges in finance as identified by \cite{KHAN2024e24890} still persist. We also discuss carbon emission comparison of models in Appendix \ref{sec:env}. 

\section{Market Analysis}
\label{sec:market_analysis}
\subsection{Experiment Setup}
\paragraph{Construction of the Optimism Measure}
We use our weak-supervision model to label all the financial numeric sentences in the analyst reports and earnings calls as in-claim or out-of-claim. We then filter the sentences and only keep in-claim sentences to evaluate predictions.

We further label each in-claim sentence as `positive', `negative', or `neutral' using the \href{https://huggingface.co/ipuneetrathore/bert-base-cased-finetuned-finBERT}{fine-tuned} sentiment analysis model specifically for the financial domain. The model is fine-tuned for financial sentiment analysis using the pre-trained FinBERT \citep{araci2019finbert}. We then use labeled sentences in each document to generate a document-level measure of analyst optimism for document $i$ using the following formula:
\begin{equation}
    \footnotesize
    \text{Optimism}_{i} = 100 \times  \frac{\text{Pos. In-claim}_i - \text{Neg. In-claim}_i}{\text{Total Sentences}_{i}}
\end{equation}

where $\text{Pos. In-claim}_{i}$ and $\text{Neg. In-claim}_{i}$ are the number of positive and negative in-claim sentences respectively in document $i$ after the filter, and $\text{Total Sentences}_{i}$ is the total number of sentences in the document. 

\paragraph{Empirical Specification}
We use the following empirical specification for market analysis. 

\begin{equation}
\footnotesize
Y_{i,t} = \alpha + \beta \times \text{Optimism}_{i,t} + \epsilon_{i,t}
    \label{eq:basic_reg}
\end{equation}

Here $Y_{i,t}$ is the outcome variable of interest for firm $i$ at time $t$, $\alpha$ is a constant term, and $\epsilon_{i,t}$ is an error term. The coefficient ($\beta$) will help us understand the influence of $\text{Optimism}_{i,t}$ on the outcome variable ($Y_{i,t}$). 

\begin{table}[ht]
\centering
\footnotesize
\begin{tabular}{lccc}
\toprule
\textbf{Outcome ($Y$)} & \textbf{Constant ($\alpha$)} & \textbf{Beta ($\beta$)} \\
\midrule
Earn. Surp. & 0.1744 *** & -1.9883 ***   \\
\midrule

CAR [+2, +30] &  0.9548  *** &  -34.5749 *** \\

CAR [+2, +60] & 0.8559 ** &  -54.335 ***  \\
\bottomrule
\end{tabular}
\caption{Market analysis result based on the empirical regression. *, **, and *** indicate significance at the 10\%, 5\%, and 1\% levels, respectively.}
\label{tb:results_reg}
\end{table}

\subsection{Post Earnings Prediction}

We examine the relation between optimism in analyst reports for a company in a specific quarter and its effect on earnings. Using earnings-based metrics, we perform a regression as per Eq~\ref{eq:basic_reg} using earnings call transcripts and analyst report data. For quarters with multiple reports on one stock, we aggregate sentences and claims to compute $\text{Optimism}_i$.

\paragraph{Earnings Surprise (\%)}

The Earning Surprise (\%) is calculated by subtracting the median EPS (in the last 90 days) from the actual EPS. The difference is scaled by the stock price at the end of the quarter and multiplied by 100.  This method aligns with \citet{chava2022measuring}. 

The Earnings Surprise (\%) is set as the outcome variable ($Y_{i,t}$). The results in Table~\ref{tb:results_reg} show a significant link between optimism and the Earnings Surprise (\%). A negative $\beta$ coefficient indicates that with every unit rise in optimism in analyst reports, the Earnings Surprise (\%) drops. This implies that heightened optimism in reports often leads to the actual EPS underperforming expectations. This "false optimism" aligns with previous studies like \citep{coleman2021human}, highlighting analysts' tendency to overestimate firm performance.

\paragraph{Cumulative Abnormal Returns}

We further aim to explore the influence of optimism in analyst reports on the magnitude of cumulative abnormal return (CAR) post-earnings. CAR for a firm represents the total daily abnormal stock return in the period after a specific event, in our context, the firm's earnings conference call.

We analyze two CAR time frames. CAR[+2, +30] is the cumulative abnormal for the [+2,+30] trading day window post-earnings call, as determined by \citet{chava2022measuring}. The same methodology is used to calculate CAR[+2, +60] as well. 

Table~\ref{tb:results_reg} shows that greater optimism in analyst reports corresponds with a larger decline in CAR. This emphasizes the 'false optimism' trend in reports, where increased optimism leads to greater discrepancies from actual outcomes, leading to a larger negative cumulative abnormal return.

The prevailing notion in finance literature is that analysts are overly optimistic. While \citet{francis1993analysts} and \citet{barber2007comparing} believe this bias helps maintain good ties with corporate insiders, \citet{michaely1999conflict} sees it as a means for personal financial gains. Recently, \citet{brown2022analysts} found that analysts favor firms with attributes like high debt or fluctuating earnings. This suggests such firms might exaggerate earnings, potentially through manipulation. Our market analysis aligning with these theories reinforces our method's accuracy and the financial relevance of our study. Furthermore, \citet{bhojraj09} shows that simply exceeding or failing to meet analyst expectations under certain conditions can lead to unique post-earnings characteristics for a company.

\subsection{Predictive Power of Optimism}
To highlight a usage of Optimism for making trading predictions, we employ a simple ``trading strategy''. We utilize analyst reports from 2017-2019 as a training set to identify the average positive bias in the "optimism" measure. To adjust for the bias in our test set, the 2020 analyst reports, we subtract the mean bias from the optimism score for each company, correcting for the inherent positive bias. The division of the dataset into training and testing phases is crucial to avoid look-ahead bias in calculating mean optimism. After adjusting the optimism measure in the test dataset, we implement a straightforward investment strategy: short selling companies with a positive adjusted optimism score and buying shares of companies with a negative adjusted optimism score. This approach is based on the rationale of investing in companies with \textbf{overly} pessimistic sentiment and divesting from those with \textbf{overly} optimistic sentiment. We use Earnings Surprise, CAR[+2, +30], and CAR[+2,+60] to determine the success or failure of our hypothetical trades.

\begin{figure}
    \centering
    \includegraphics[width=.95\linewidth]{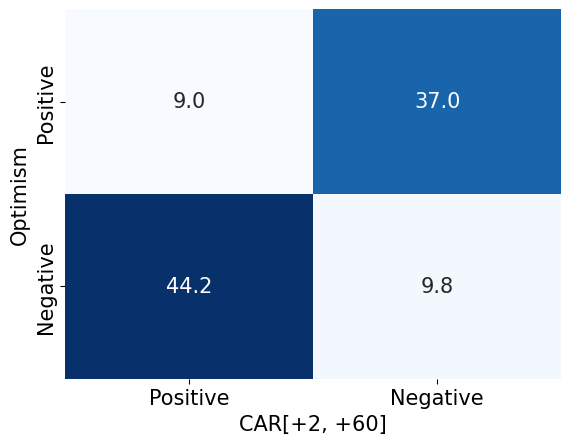}
    \caption{Normalized Confusion Matrix illustrating the percentage of trades categorized by negative or positive adjusted optimism and their corresponding CAR[+2,+60] outcomes. Each cell represents the percentage of total trades that fall within each category.}
    \label{fig:car260confusion}
\end{figure}
The confusion matrix corresponding to the results of CAR[+2,+60] are visualized in Figure \ref{fig:car260confusion}, while Earnings Surprise and CAR[+2,+30] are shown in Appendix \ref{ap:add_predi_power}. The confusion matrix shows that such a rule-based strategy achieves an approximate  81\% accuracy in correctly predicting the direction of stock movement. Additionally, the high accuracy lasting up to 60 days indicates that using optimism can effectively predict stock movements for more than just a few days, demonstrating a valuable preliminary application of such identification for the financial field.

\section{Conclusion}
Our work presents claim based labeled dataset in the English language alongside presenting a weak-supervision model with an accuracy of 93\%. Developed customized aggregation function outperforms baseline aggregation functions. We benchmark various language models and compare the performance with the weak-supervision model. We show the application of claim detection by generating a measure of optimism from the weak-supervision model. We also validate the measure by studying its applicability in predicting earnings surprise, abnormal returns, and earnings optimism. We release our models, code, and benchmark data (for earnings call transcripts only) on Hugging Face and GitHub. We also note that the trained model for claim detection can be used on other financial texts.

\section*{Limitations}
By acknowledging the following limitations, we pave the way for future research to address these areas and further enhance the understanding and applicability of our approach.

\begin{itemize}
    \item \textit{Limited Scope of Text Data}: Our analysis is restricted to analyst reports and earnings calls, excluding other potentially valuable text datasets such as related news articles and investor presentations. Incorporating these additional sources of information could provide a more comprehensive understanding of pre-earnings drifts.

    \item \textit{Exclusion of Audio and Video Features}: Our measure construction does not utilize audio or video features from earnings calls, which may contain supplementary information. 

    \item \textit{Omission of Alternative Weak-Supervision Models}: We do not explore multiple end models, such as the confidence-based sampling with contrastive loss proposed in the COSINE framework by \citet{yu2020fine}. Incorporating such alternative weak-supervision models could offer additional insights and improve the robustness of our approach.

\end{itemize}

\section*{Ethics Statement}
Our work adheres to ethical considerations, although we acknowledge certain biases and limitations in our study. We do not identify any potential risks stemming from our research; however, we recognize the presence of geographic and gender biases in our analysis.
\begin{itemize}
    \item \textit{Geographic Bias}: Our study focuses solely on publicly listed companies in the United States of America, which introduces a geographic bias. The findings may not be fully representative of global firms and markets. 
    \item \textit{Gender Bias}: We acknowledge the gender bias present in our study due to the predominant representation of male analysts, CEOs, and CFOs. 
    \item \textit{Data Ethics}: The data used in our study, derived from publicly available sources, does not raise ethical concerns. All raw data is obtained from public companies that are obligated to disclose information under the guidance of the SEC and are subject to public scrutiny. 
    \item \textit{Language Model Ethics}: The language models employed (with proper citation) in our research are publicly available and fall under license categories that permit their use for our intended purposes. While most models employed are publicly available, it is important to note that ChatGPT's prompt answers will not be made public due to licensing conditions.  We acknowledge the environmental impact of large pre-training of language models and mitigate this by limiting our work to fine-tuning existing models. 
    \item \textit{Annotation Ethics}: All annotations were performed by the authors, ensuring that no additional ethical concerns arise from the annotation process. 
    \item \textit{Hyperparameter Reporting}: In the interest of clarity and readability, we refrain from reporting the best hyperparameters found through grid search in the main paper. Instead, we will make all grid search results, including hyperparameter information, publicly available on GitHub. This transparency allows interested readers to access detailed information on our experimental setup.
    \item \textit{Publicly Available Data}: We specify the datasets that will be made publicly available and indicate the applicable licenses under which they will be shared. 
\end{itemize}

By acknowledging these ethical considerations and limitations, we strive to maintain transparency and promote responsible research practices.

\bibliography{anthology,custom}

\appendix

\section{Robustness Check} 
\label{sec:robustness}
From a data engineering perspective, there can be concern about the model design and gold data construction as the authors who designed the weak-supervision model have annotated the data. This can lead to exaggerated performance on the data, which may taint the test set. To ensure that there is no contamination issue in the weak-supervision model and it is generalizable, we get the same test dataset annotated separately by four annotators with master's degrees in Quantitative Finance. These annotators were hired by the department as Graduate Assistants based on merit and were paid a \$20 per hour salary for their work which is more than double the federal minimum wage and higher than the highest minimum wage (\$15.74 in Washington, D.C) in the USA. The rates are standard and in compliance with ethical standards. These annotators had no information about the rules/patterns used in our weak-supervision model. Each sample in the test dataset is annotated by two annotators, and we drop the observations where there is a disagreement among annotators. \footnote{There is 98.59\% agreement between two annotators.} The F1 score of the weak-supervision model on a dataset annotated by non-authors is 0.9281 which is close to a score of 0.9272 on the author-annotated dataset. We also recalculate the F1 score of the model based on the author-annotated labels after dropping observations dropped in a non-author annotated dataset. The model gives a higher mean F1 score of 0.9360 which is expected as ambiguous sentences are dropped. Overall these results show the robustness of our model on the dataset annotated by annotators who don't have knowledge of the rules used in the weak-supervision model. From here onwards, the performance is always calculated on a gold dataset created by authors.

\begin{table*}
\footnotesize
\begin{tabular}{ cp{0.23\textwidth}p{0.08\textwidth}p{0.17\textwidth}p{0.35\textwidth}}
 \toprule
 \centering \textbf{Set} & \textbf{Used to detect} & \centering \textbf{Output} & \centering \textbf{Type} &  \textbf{Keyword or phrase} \\
 \midrule
\centering 1 & High Confidence out-of-claim (Past Tense or Assertions) & \centering  -1/0 & \centering Phrase Matching &  reasons to buy:, reasons to sell:, was, were, declares quarterly dividend, last earnings report, recorded \\
 \centering 2 & Low Confidence in-claim & \centering  1/0 & \centering Phrase Matching &  earnings guidance to, touted to, entitle to\\
\centering 3 &  High Confidence in-claim & \centering 2/0 & \centering Lemmatized Word matching  & expect, anticipate, predict, forecast, envision, contemplate\\
 \centering 4 &  High Confidence in-claim & \centering 2/0 & \centering POS Tag for word ``project" & VBN, VB, VBD, VBG, VBP, VBZ\\
 \centering 5 &  High Confidence in-claim & \centering  2/0 & \centering Phrase Matching &  to be, likely to, on track to, intends to, aims to, to incur, pegged at\\
 \bottomrule
\end{tabular}
\caption{Labeling Functions used in weak-supervision model. SpaCy Lemmatizer has been used for labeling functions involving lemmatized word matching.}
\label{labeling_functions}
\end{table*}

\section{Labeling Functions Methodology}
\label{sec:ap_labeling_function}
The following illustrates the methodology adopted by us while choosing the rules to define the weak-supervision mode. All rules were acknowledged post detailed analysis of sample documents distributed over sector and time :
\begin{enumerate}
    \item Certain phrases such as "reasons to buy", "reasons to sell" or the presence of words which are indicative of past tense such as "was", "were" are characteristic of out-of-claim sentences, since they indicated either facts or events which happened in the past. Examples are given in the set 1 of Table \ref{labeling_functions}. 
    \item Phrases often provided definitive information about a given sentence in a document and in most cases they had a fairly consistent linguistic composition. Examples are given in the set 2 of Table \ref{labeling_functions}. 
   \item In a bid to capture the effect of a few other verb forms indicative of a probabilistic event, we also chose to look at its lemmatized form to reduce inflectional usage and use the base token for a more holistic evaluation over multiple usage formats. Examples are given in the set 3 of Table \ref{labeling_functions}.  
   \item POS tags were also derived for "project" as a word wherever present. This was done to segregate its usage as a verb. Its usage as a verb was usually observed to be adopted while making claims or predictions. Examples are given in the set 4 of Table \ref{labeling_functions}. 
   \item The alternate adoption of phrase matching was to identify in-claim sentences. This mostly consisted of a verb form indicative of a probabilistic event (eg: likely, intends) coupled with a preposition (usually "to" or "at"). Based on the ambiguity of the resulting phrase they were either categorised as a high-confidence claim or a low-confidence one. Examples are given in the set 5 of Table \ref{labeling_functions}. 
\end{enumerate}

\section{Additional Models}
\label{ap:model_details}
\subsection{Generative LLMs}
To understand the capabilities of current state-of-the-art  (SOTA) generative LLMs' in a zero-shot and few-shot manner, we add ChatGPT\footnote{\url{https://chat.openai.com/}} performance benchmark in our study. We use the "gpt-3.5-turbo-0613" model with 200 max tokens for output, and a 0.0 temperature value. The ChatGPT API was accessed on Feb 2nd, 2024. In a recent article, \citet{rogers-etal-2023-closed} made a case for why closed models like ChatGPT make bad baselines. In order to understand where SOTA open-source LLMs stand in comparison to ChatGPT and fine-tuned models, we also test the Falcon-7B-Instruct \citep{falcon40b} and "Llama-2-70B-chat" \citep{touvron2023llama} models.  The prompt templates are provided in Table \ref{tb:prompts}. All our prompting was done in consistency with reputable resources, such as the “Prompt Engineering Guide” \footnote{\url{https://www.promptingguide.ai/}}. We also test the model with zero-shot and six-shot. The six-shot prompting consists of 3 `in-claim' examples and 3 `out-of-claim' examples. 

\begin{table*}[h!]
\centering
\begin{tabular}{cp{12.5cm}}
\toprule
\textbf{Prompt Name} & \textbf{Description} \\ \hline
\textbf{Zero-shot} & \texttt{Discard all the previous instructions. Behave like you are an expert sentence classifier. Classify the following sentence into either `INCLAIM' or `OUTOFCLAIM'. `INCLAIM' refers to predictions or expectations about financial outcomes. `OUTOFCLAIM' refers to sentences that provide numerical information or established facts about past financial events. For each classification, `INCLAIM' can be thought of as `financial forecasts', and `OUTOFCLAIM' as `established financials'. Now, for the following sentence provide the label in the first line and provide a short explanation in the second line. The sentence: \{sentence\}} \\ 
\midrule
\textbf{Few-shot} & \texttt{Discard all the previous instructions. Behave like you are an expert sentence classifier. Classify the following sentence into either `INCLAIM' or `OUTOFCLAIM'. `INCLAIM' refers to predictions or expectations about financial outcomes. `OUTOFCLAIM' refers to sentences that provide numerical information or established facts about past financial events. For each classification, `INCLAIM' can be thought of as `financial forecasts', and `OUTOFCLAIM' as `established financials'. Here are a few examples: \ Example 1: free cash flow of \$2.3 billion was up 10.5\%, benefiting from the positive year-over-year change in net working capital due to covid at both nbcu and sky, half of which resulted from the timing of when sports rights payments were made versus when sports actually aired and half of which resulted from a slower ramp in content production. // The sentence is OUTOFCLAIM \ Example 2: we've also used our scale of more than 15,000 combined stores to drive merchandise cost savings exceeding \$70 million. // The sentence is OUTOFCLAIM \ Example 3: consolidated total capital was \$2.9 billion for the quarter. // The sentence is OUTOFCLAIM \ Example 4: third, as a result of the continued strength of the u.s. dollar, we are now factoring in an incremental fx headwind of \$175 million across q3 and q4 revenue. // The sentence is INCLAIM \ Example 5: though early, we are planning our business based on the expectation of cy '23 wfe declining approximately 20\% based on increasing global macroeconomic concerns and recent public statements from several customers, particularly in memory, and the impact of the new u.s. government regulations on native china investment. // The sentence is INCLAIM \ Example 6: we expect revenue growth to be in the range of 5.5\% to 6.5\% year on year. // The sentence is INCLAIM \ Now, for the following sentence provide the label in the first line and provide a short explanation in the second line. The sentence: \{sentence\}} \\ 
\bottomrule
\end{tabular}
\caption{Prompts used for zero-shot and few-shot inference. }
\label{tb:prompts}
\end{table*}

\subsection{Bi-LSTM}
In the realm of text classification problems, Long Short-Term Memory (LSTM) was a popular recurrent neural network architecture \citep{hochreiter1997long}. An enhanced approach to LSTM is the Bidirectional LSTM (Bi-LSTM), which processes input in both directions \citep{schuster1997bidirectional}. In order to assess the efficacy of Recurrent Neural Networks (RNNs) in claim detection, we employ the Bi-LSTM model on the datasets we have developed. Instead of training it from scratch, we initialize the embedding layer of the Bi-LSTM using 300-dimensional GloVe embeddings trained using Common Crawl \citep{glove}. Here we perform the task of sequence classification while minimizing the cross-entropy loss. We employ a grid search approach to identify the optimal hyperparameters for each model, considering four different learning rates (1e-4, 1e-5, 1e-6, 1e-7) and four different batch sizes (32, 16, 8, 4). In our training process, we employ a maximum of 100 epochs, incorporating early stopping criteria. In cases where the validation F1 score does not exhibit an improvement of greater than or equal to 1e-2 over the subsequent 7 epochs, we designate the previously saved best model as the final fine-tuned model. 

\subsection{PLMs}
In order to establish a performance benchmark, our study encompasses a range of transformer-based \citep{vaswani2017attention} models of varying sizes. For the small models, we employ BERT \citep{devlin2018bert}, FinBERT \citep{finbert}, FLANG-BERT \citep{shah-etal-2022-flue}, and RoBERTa \citep{roberta}. Within the category of large models, we incorporate BERT-large \citep{devlin2018bert} and RoBERTa-large \citep{roberta}. To avoid over-fitting on financial text, we refrain from conducting any pre-training on these models prior to fine-tuning. Here we perform the task of sequence classification while minimizing the cross-entropy loss. For PLMs, we employ grid-search, fine-tuning, and early stopping similar to what we used for Bi-LSTM. The experiments are conducted using PyTorch \citep{pytorch} on an NVIDIA RTX A6000 GPU. Each model is initialized with the pre-trained version from the Transformers library provided by Huggingface \citep{huggingface}.

\section{Ablation: Number of Labeling Functions} 
\label{ap:number_lfs}
\begin{figure}[ht]
\centering
\includegraphics[width=0.5\textwidth]{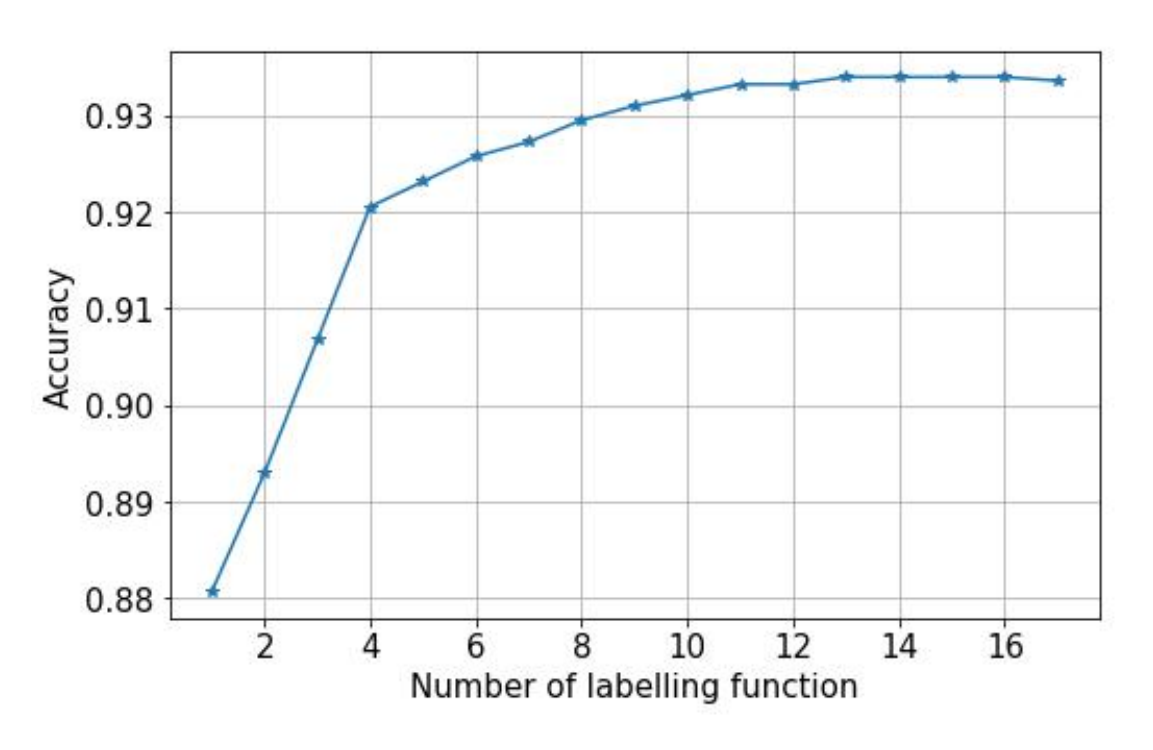}
  \caption{Accuracy v/s Number of labeling functions. Note: This is accuracy, not F1 score. }
  \label{accuracy_curve}
\end{figure}

Figure~\ref{accuracy_curve}, shows how the accuracy of the model changes depending on the number of labeling functions. For this plot, we initially computed the contribution of each labeling function (Table \ref{labeling_functions}, High confidence and Low Confidence in-claim)
towards the detection of in-claim sentences and then considered the addition of new labeling function at each step to ensure the steepest ascent to saturation. At each step, in addition to one new labeling function, all labeling functions present in Table \ref{labeling_functions} for Past Tense and Assertions, were also used. They either abstain or classify sentences as out-of-claim and help improve the classification of out-of-claim sentences. From the plot, we can notice that after around thirteen labeling functions, the addition of new labeling functions does not produce any change in the accuracy. In fact, increasing labeling functions thereafter leads to a minor decrease in accuracy. This suggests that we can effectively capture the required trends for classification in this setting with thirteen labeling functions. 

\section{Environmental Impact}
\label{sec:env}
\begin{figure}
    \centering
    \includegraphics[width=\linewidth]{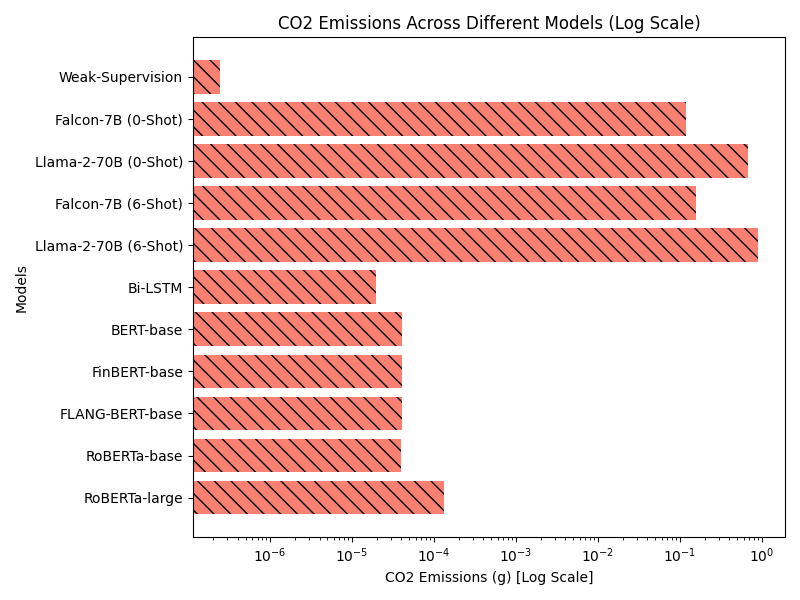}
    \caption{This bar chart compares the CO2 emissions (log scale) of various models relative to the weak-supervision model.}
    \label{fig:environment}
\end{figure}

Our investigation extends beyond just performance metrics, embracing a conscientious approach towards the environmental implications of AI usage. 
To ensure a standardized and rigorous assessment of CO2e, we drew upon the methodology outlined by \citet{lannelongue2021green} and utilized the Green Algorithms calculator\footnote{\url{https://calculator.green-algorithms.org/}}. The value of CO2e are reported in Figure \ref{fig:environment}. This dual focus on minimizing latency and CO2e without compromising performance highlights our commitment to advancing sustainable and efficient AI technologies in sectors where both are of paramount importance, such as finance. The CO2 emissions (CO2e) associated with the inference phase of these models are particularly telling, with our WS model not only leading in latency but also in sustainability, registering the lowest CO2e among all models reviewed. This underscores the viability of employing AI in environments where both speed and environmental responsibility are valued. In contrast, models such as Llama-70B, despite their performance coming close to our model, incur significantly higher (more than a million times larger) CO2e due to their reliance on extensive GPU resources. 

\begin{table*}[htbp!]
\centering
\footnotesize
\begin{tabular}{lccccc} 
\toprule
 & & \multicolumn{1}{c}{\textbf{ES (\%)}}  & \multicolumn{1}{c}{\textbf{CAR [+2,+30]}} & \multicolumn{1}{c}{\textbf{CAR [+2, +60]}} \\
\midrule 
\textbf{Sentence Type/Subset}& \textbf{Average Sentences} &   \textbf{Adj. $\beta$} &  \textbf{Adj. $\beta$}& \textbf{Adj. $\beta$}\\ 
\midrule 
\textit{Unfiltered} & 98 & -0.054***   & -0.02** & -.03*** \\ 

\textit{Numeric} & 26 &   -0.28*** & -.06***  & -.09*** \\ 

\textit{Numeric Financial} & 21.6 &  -0.29***   & -.07*** & -.11*** \\ 

\textit{Numeric Financial In-claim} & 3.7 &  -1.51*** & -.26*** & -.41*** \\ 
\bottomrule 
\end{tabular}  
\caption{Ablation on market analysis, highlighting the importance and information density of ``in-claim'' sentences. *, **, and *** indicate significance at the 10\%, 5\%, and 1\% levels, respectively.}
\label{tb:ablation_market}
\end{table*}

\section{Ablation Study: Market Analysis}
\label{sec:ablation_market}
To understand the influence of ``in-claim'' sentences on market sentiment, we introduce the optimism measure in section \ref{sec:market_analysis}, outlining its implications. In this section, we carry out an ablation study to better understand the impact of ``in-claim'' sentences. As such, we compute the optimism score for four sentence subsets: Unfiltered, Numerical, Numerical Financial, and Numerical Financial ``In-claim'' sentences for each file. For example, the optimism score for a subset of Numerical sentences for document $i$ is given by:

\begin{equation*}
    \scriptsize
    \text{Optimism (Numerical)}_{i} = 100 \times \frac{\text{Pos. Numerical}_i - \text{Neg. Numerical}_i}{\text{Total Sentences}_{i}}
\end{equation*}

We standard normalize these scores for uniform comparison by deducting their mean and dividing by the standard deviation. As the beta coefficient lacks full context, to factor in the size of the sentence subset, we adjusted each coefficient by the average sentence count, terming it as the adjusted beta. This illustrates the information density in each filtered sentence set. When examining the Earnings Surprise (\%) columns of Table \ref{tb:ablation_market} the Adjusted Beta for Earnings Surprise increases, implying that a mere average of 3.7 ``in-claim'' sentences holds crucial information. This highlights the high information density of our filtered sentences. While we aren't dismissing the importance of other sentences, our analysis reveals that the ones we've extracted are the most informative on a per-sentence basis. 

\section{Predictive Power of Optimism (Earnings Suprise and CAR[+2,+30])}
\label{ap:add_predi_power}

\begin{figure}[ht]
    \centering
    \includegraphics[width=1\linewidth]{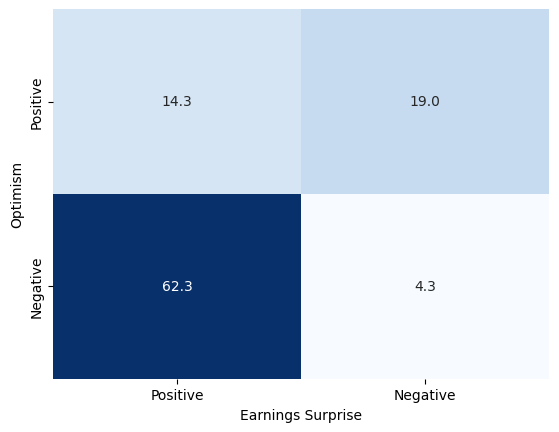}
    \caption{Percentage of trades categorized by negative or
positive adjusted optimism and their corresponding
Earnings Surprise outcomes.}
    \label{fig:esconfusion}
\end{figure}

\begin{figure}[ht]
    \centering
    \includegraphics[width=1\linewidth]{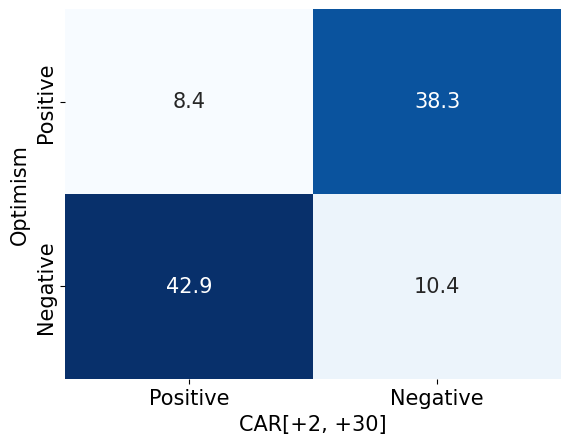}
    \caption{Percentage of trades categorized by negative or
positive adjusted optimism and their corresponding
CAR[+2,+30] outcomes.}
    \label{fig:car230confusion}
\end{figure}

Figure \ref{fig:esconfusion} and \ref{fig:car230confusion} show the results of making trades based on a positive or negative adjusted optimism in terms of the respective performance of the company.

\end{document}